\documentclass[10pt,twocolumn,letterpaper]{article}

\usepackage[pagenumbers]{cvpr}

\usepackage{colortbl}
\newcommand{\modelname}{\textsc{Planet}\xspace}
\newcommand{\modelnamewithplanet}{%
  \textsc{Planet}\,\raisebox{-0.28ex}{%
    \includegraphics[height=1.35em]{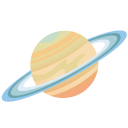}}%
}

\newcommand{\cmark}{\ensuremath{\checkmark}}
\newcommand{\xmark}{\ensuremath{\times}}

\definecolor{cvprblue}{rgb}{0.21,0.49,0.74}
\usepackage[breaklinks,colorlinks,allcolors=cvprblue]{hyperref}
\hypersetup{
pdftitle={Beyond the Image Plane: World-Grounded Queries for Multi-Object Tracking},
pdfauthor={Orcun Cetintas, Guillem Bras\'o, Tim Meinhardt, Laura Leal-Taix\'e}
}

\title{Beyond the Image Plane: World-Grounded Queries for Multi-Object Tracking}

\author{
Orcun Cetintas$^{1,2}$ \quad
Guillem Bras\'o$^{1}$ \quad
Tim Meinhardt$^{1}$ \quad
Laura Leal-Taix\'e$^{1}$\\
$^{1}$NVIDIA \qquad $^{2}$Technical University of Munich
}

\begin{document}
\maketitle
\begin{abstract}
Monocular videos record 3D scenes as sequences of 2D image-plane projections, obscuring depth and spatial relationships.
Multi-object trackers localize and associate objects primarily using appearance and geometry observed only in the image plane, inheriting these ambiguities.
To address this limitation, we introduce \modelname, an end-to-end multi-object tracker designed to move beyond the image plane.
As an enabling step, we lift existing 2D tracking datasets into 3D.
We then form \emph{world-grounded queries} by embedding reconstructed 3D scene geometry into the features and positional encodings used during query formation.
An auxiliary 3D location prediction task further encourages the queries to encode object positions during training.
A complementary dual-resolution temporal memory preserves this evidence across longer temporal gaps.
As a result, \modelname achieves state-of-the-art performance across three diverse benchmarks.
\end{abstract}

\section{Introduction}
\label{sec:introduction}

Multi-object tracking aims to localize multiple objects in a video while preserving their identities over time.
Tracking-by-detection has long served as the dominant paradigm for multi-object tracking, decomposing the problem into object detection followed by a separately designed data-association stage~\cite{zhang2022bytetrack,braso2020learning,zhou2020tracking,cetintas2023unifying,lv2024diffmot}.
More recently, end-to-end trackers have begun to replace this separation, primarily through Transformer-based architectures that connect detection and identity reasoning with object queries~\cite{meinhardt2022trackformer,zeng2022motr,gao2023memotr,sun2020transtrack}.
Although these models commonly operate with limited temporal context~\cite{meinhardt2022trackformer,zeng2022motr,gao2023memotr,segu2025samba}, their query-based formulation brings the two tasks into a shared trainable model and provides a flexible basis for improving the representations used by both.
Despite their different formulations, both families face a common set of challenges, including occlusion, camera motion, object motion, viewpoint changes, and similar object appearances~\cite{sun2022dancetrack}.

A central source of these difficulties is the mismatch between the physical scene and its visual observation: objects occupy and move through a 3D world, while a camera records their projection onto a 2D image plane.
This projection obscures depth and spatial configuration, allowing distinct objects to overlap in the image and making physical motion ambiguous in 2D.
Trackers reason primarily from appearance and image-plane position or motion, and therefore inherit these geometric ambiguities.
3D information offers a natural way to reduce these ambiguities.
A line of prior work shows that 3D cues can aid occlusion reasoning~\cite{wojek2011monocular}, camera-motion compensation~\cite{hu2019joint}, trajectory forecasting~\cite{dendorfer2022quo}, and association~\cite{osep2017combined}.
However, these approaches are generally tied to category-specific 3D models or handcrafted downstream association, leaving the integration of 3D information into multi-object trackers underexplored.

Until recently, recovering sufficiently complete scene geometry and camera information from monocular video relied largely on costly optimization-based pipelines that could struggle in challenging conditions.
Recent feed-forward geometry models, including VGGT~\cite{wang2025vggt} and Depth Anything 3~\cite{lin2025depth}, make this process more efficient and robust by estimating dense 3D scene structure and camera parameters from image sequences with a fully learnable model.
These new capabilities create an opportunity to embed estimated 3D scene information directly into 2D tracking models, enriching the internal representations used for detection and association.
This geometric grounding provides an opportunity for trackers to move beyond the image plane and reason about the underlying scene.

Motivated by moving beyond the image \underline{\textbf{\textit{PLANE}}} for \underline{\textbf{\textit{T}}}racking, we introduce \textbf{\textit{PLANET}}\,\raisebox{-0.28ex}{\includegraphics[height=1.35em]{fig/emoji/noto_ringed_planet.png}}, an end-to-end multi-object tracking model that incorporates 3D scene information.
Our approach proceeds in two stages.
First, using Depth Anything 3~\cite{lin2025depth}, we lift three existing 2D tracking datasets into 3D: DanceTrack~\cite{sun2022dancetrack}, SportsMOT~\cite{cui2023sportsmot}, and BFT~\cite{zheng2024nettrack}.
In doing so, we recover dense world-coordinate point maps and associate each annotated 2D bounding box with a pseudo-3D object center.
Second, \modelname incorporates this lifted scene information into the tracker through 3D embeddings and auxiliary 3D location prediction.
The 3D embeddings expose detector queries to the reconstructed scene geometry, while auxiliary 3D location prediction encourages each matched query to encode its object's 3D location.
Together, these components form \emph{world-grounded queries} that enrich the representation shared by detection and association with both scene-level geometry and object-level 3D supervision.

World-grounded queries can provide stronger spatial evidence, but this evidence can support association after an occlusion only if it remains available to the tracker.
However, existing end-to-end trackers commonly operate within a limited temporal horizon because retaining and processing a growing query history is computationally costly~\cite{meinhardt2022trackformer,zeng2022motr,gao2023memotr,segu2025samba}.
To address this, \modelname introduces \emph{dual-resolution temporal memory}, which preserves recent observations densely while sampling older observations more sparsely.
Under the same context budget, \modelname expands the accessible temporal horizon from 30 frames in MOTIP~\cite{gao2025multiple} to approximately 200 frames.

We evaluate \modelname on DanceTrack~\cite{sun2022dancetrack}, SportsMOT~\cite{cui2023sportsmot}, and BFT~\cite{zheng2024nettrack}, which collectively cover challenging human motion, sports scenes, and dense non-human motion.
Across all three benchmarks, \modelname achieves state-of-the-art performance.
Controlled ablations further examine how 3D embeddings, auxiliary 3D location prediction, and dual-resolution temporal memory contribute to tracking performance.
Together, these results highlight 3D information as a valuable source of evidence for multi-object tracking.
Upon acceptance, we will release our code, trained models, and the 3D-enriched versions of the datasets.

\section{Related Work}
\label{sec:related-work}

\paragraph{Multi-Object Tracking.}
Tracking-by-detection decomposes multi-object tracking into per-frame object detection followed by a separately designed data-association stage.
This stage commonly links detections using appearance descriptors~\cite{wojke2017simple}, image-plane motion and spatial overlap~\cite{bewley2016simple,cao2023observationcentric}, detection confidence~\cite{zhang2022bytetrack}, or learned motion models~\cite{lv2024diffmot}.

End-to-end approaches instead optimize detection and identity association jointly.
Early Transformer-based trackers such as TransTrack~\cite{sun2020transtrack} and TrackFormer~\cite{meinhardt2022trackformer} connect the two tasks through learned object and track queries; subsequent methods follow and improve this paradigm~\cite{zeng2022motr,gao2023memotr,yan2025comot,segu2025samba}.
MOTIP provides a particularly direct formulation in which detection and ID prediction operate on the same detector-query representations through their respective prediction heads~\cite{gao2025multiple}.

Despite this progress, end-to-end trackers also face a temporal-context bottleneck.
Extending the attention horizon requires retaining and processing more query observations, which increases computational cost and distributes attention over an increasingly long and redundant history.
Consequently, identity-relevant evidence from distant frames can become diluted.
Reflecting this trade-off, many recent end-to-end Transformer trackers use short training contexts of only two to ten frames~\cite{meinhardt2022trackformer,zeng2022motr,zhang2023motrv2,gao2023memotr,segu2025samba}.
This creates a need to extend temporal coverage while retaining the most useful recent and historical evidence.

Building on MOTIP's shared-query formulation, \modelname grounds queries in reconstructed 3D geometry, allowing the same representation to inform detection and association.
A complementary dual-resolution memory preserves dense short-term and sparse long-term context, extending the temporal support available for identity prediction under a fixed context budget.

\paragraph{3D Cues for Monocular Tracking.}
Previous work incorporates 3D cues through category-specific representations, including pedestrian visibility reasoning~\cite{wojek2011monocular}, world-space trajectories~\cite{hu2019joint}, and 3D human models~\cite{rajasegaran2022tracking}.
Such representations aid occlusion reasoning and camera-motion compensation, but rely on category-specific models or assumptions about ego-motion.

Occlusion-focused approaches instead pursue object permanence through recurrent invisible-object supervision~\cite{tokmakov2021learning}, depth-constrained Kalman forecasting~\cite{khurana2021detecting}, or bird's-eye-view trajectory prediction~\cite{dendorfer2022quo}.
Here, geometry is handled by separate forecasting, search, or re-identification modules and remains independent of the tracker's internal object representation.

Geometry has also been incorporated into conventional 2D tracking pipelines during downstream association, through handcrafted geometric and appearance costs~\cite{sharma2018beyond}, coupled 2D--3D filtering~\cite{osep2017combined}, depth and motion heuristics~\cite{pang2021quasidense}, or learned depth--motion--appearance likelihoods~\cite{mancusi2023trackflow}.
Despite their different fusion strategies, these methods introduce geometry only after the object representations have been formed, leaving the overall tracking system as a multi-stage pipeline rather than a unified model.

In contrast, \modelname remains an end-to-end monocular tracker but introduces reconstructed scene geometry directly into query formation.
Dense point maps and auxiliary 3D center localization produce world-grounded queries before downstream association, allowing the same representation to benefit both detection and identity reasoning.
This avoids specialized category-specific models, external forecasting, and handcrafted cues.

\section{Methodology}
\label{sec:methodology}

\begin{figure}[t]
    \centering
    \includegraphics[width=\columnwidth]{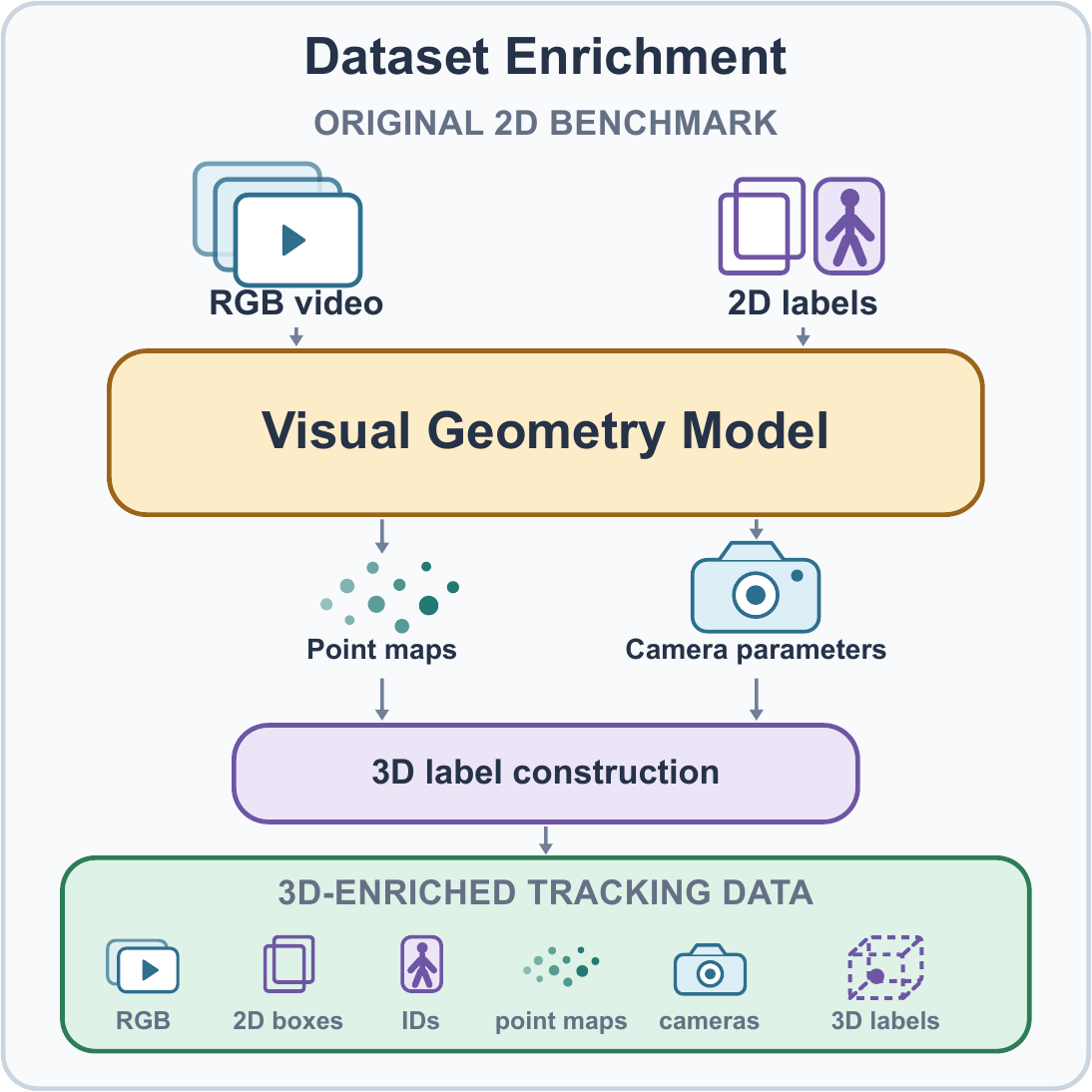}
    \caption{\textbf{Lifting 2D tracking benchmarks into 3D.} We enrich existing 2D tracking benchmarks with reconstructed scene geometry and 3D object labels.}
    \label{fig:dataset-enrichment}
\end{figure}

\begin{figure*}[t]
    \centering
    \includegraphics[width=\textwidth]{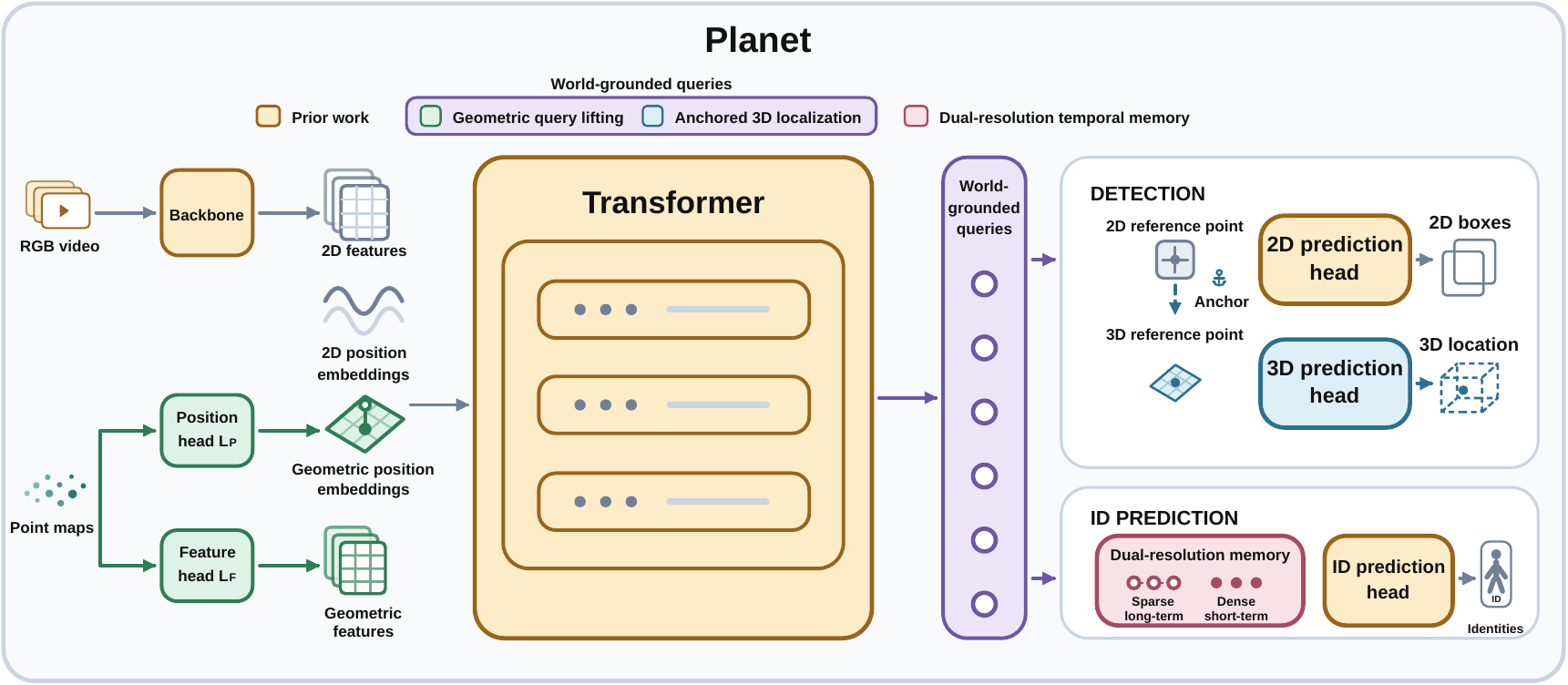}
    \caption{\textbf{\modelname overview.} Geometric query lifting conditions query formation on dense scene geometry, while anchored 3D localization encourages matched queries to encode their 3D locations. Together, they form world-grounded queries. Dual-resolution temporal memory extends the context used for identity prediction.}
    \label{fig:pipeline}
\end{figure*}

\subsection{Preliminaries}
\label{sec:preliminaries}

We build \modelname on MOTIP~\cite{gao2025multiple}, which connects a query-based DETR detector to an ID prediction module within a single end-to-end tracking model.
MOTIP casts association as $(K+1)$-way in-context classification over a learnable ID dictionary, and its prediction head operates directly on the detector queries.
This formulation makes detection and identity prediction jointly trainable within a single model.

Formally, given a frame $I_t$, the detector produces a set of object-level query embeddings $\mathcal{Q}_t=\{\mathbf{q}_t^i\}_{i=1}^{N_t}$, with $\mathbf{q}_t^i\in\mathbb{R}^{d}$.
Each query predicts a 2D bounding box $\hat{\mathbf{b}}_t^i\in\mathbb{R}^{4}$ and an identity label $\hat{y}_t^i\in\mathcal{Y}$, yielding the frame-level output $\widehat{\mathcal{O}}_t=\{(\hat{\mathbf{b}}_t^i,\hat{y}_t^i)\}_{i=1}^{N_t}$.
Predictions with the same identity are linked across frames to form a trajectory.

With this formulation, bounding-box prediction and identity association operate on the same queries and therefore meet in a shared object representation.
Improving what the queries encode can therefore benefit both tasks simultaneously.
\modelname acts at precisely this interface by grounding the queries in estimated 3D scene structure.
The resulting queries describe not only how an object appears in the image, but also where it lies within the reconstructed scene, giving the tracker a richer basis for localization and identity reasoning.

\subsection{Method Overview}
\label{sec:method-overview}

Building on the observation that image-plane trajectories arise from objects moving through a 3D world, we bring estimated scene geometry into the query representations of a 2D tracker.
As an enabling first stage, we use the geometry foundation model Depth Anything 3~\cite{lin2025depth} to lift existing monocular tracking datasets into 3D by recovering dense world-coordinate point maps.
Since these maps do not directly provide object-level information, we introduce a dedicated assignment procedure that associates each annotated 2D box with a pseudo-3D label representing its estimated center in the reconstructed scene (\cref{fig:dataset-enrichment}).

The point maps and pseudo-3D labels serve complementary roles in \modelname.
Through \emph{geometric query lifting}, dense point maps introduce scene-level geometric evidence into the model's queries.
Through \emph{anchored 3D localization}, matched queries predict 3D object centers as an auxiliary task and learn object-level 3D information.
Geometric query lifting and anchored 3D localization address complementary aspects of the problem and together lead to the learning of our \emph{world-grounded queries} (\cref{fig:pipeline}).

However, a recurring constraint in transformer-based end-to-end tracking is the limited temporal context available to the identity model.
\modelname therefore introduces \emph{dual-resolution temporal memory}, which pairs complementary training and inference strategies to expand the temporal horizon available to the identity model.
Under this formulation, temporally varied training exposes the decoder to observations separated by different intervals.
At inference, the memory retains short-term context densely while sampling long-term history more sparsely within a fixed context budget.
This design extends the available temporal horizon from 30 frames in MOTIP to approximately 200 frames in \modelname, allowing world-grounded evidence to inform association across substantially longer gaps.

\subsection{Lifting 2D Tracking Datasets into 3D}

Existing tracking benchmarks provide monocular videos with 2D boxes and identities.
To lift these datasets into 3D, we apply DA3-Streaming, a long-sequence extension of Depth Anything 3~\cite{lin2025depth} inspired by VGGT-Long~\cite{deng2025vggt}.
At a high level, long sequences are processed with overlapping windows to obtain a sequence-level coordinate system, while loop closure limits the drift accumulated over time.
The aligned depth, camera intrinsics, and camera poses are then back-projected into dense world-coordinate point maps.

To complement the dense scene geometry with object-level labels, we assign each annotated 2D box an estimated 3D location.
For each box, we form a candidate annotation by averaging the local point cloud.
Candidates from fully visible boxes are accepted directly, while occluded boxes are handled conservatively using temporal depth, current-frame depth, and image-plane position to identify the visible foreground object.
We also improve consistency along each identity trajectory using its trajectory history.
Estimates that remain uncertain are marked invalid and excluded from 3D supervision.
The resulting 3D annotations associate each 2D box with an estimated 3D center location, without specifying its dimensions or orientation.
We also obtain camera parameters, including intrinsics and poses, although \modelname does not use them directly.
We intend to release the lifted datasets, including all additional annotations and reconstructed scene information, to support future research within and beyond tracking.

\subsection{World-Grounded Queries}
\modelname forms world-grounded queries through two complementary mechanisms: \emph{geometric query lifting} and \emph{anchored 3D localization}.
Geometric query lifting conditions query formation on dense scene geometry, while anchored 3D localization encourages the resulting queries to encode the 3D positions of their matched objects.

\subsubsection{Geometric Query Lifting}
The aligned point map $\mathbf{P}$ describes the reconstructed scene geometry.
We process $\mathbf{P}$ using two separate learnable heads: a \emph{geometric feature head} $L_F$ and a \emph{geometric position head} $L_P$.
These lifting heads produce geometric feature embeddings $\mathbf{F}_{\mathrm{3D}}=L_F(\mathbf{P})$ and geometric position embeddings $\mathbf{E}_{\mathrm{3D}}=L_P(\mathbf{P})$, respectively.
The geometric feature embeddings are added residually to the visual features $\mathbf{F}_{\mathrm{2D}}$ produced by the backbone, $\widetilde{\mathbf{F}}=\mathbf{F}_{\mathrm{2D}}+\mathbf{F}_{\mathrm{3D}}$, allowing local 3D structure to complement the appearance representation inherited from the pretrained backbone.
In parallel, the geometric position embeddings are combined with the sine-cosine 2D positional encoding $\mathbf{E}_{\mathrm{2D}}$ through a learned fusion head, $\widetilde{\mathbf{E}}=\mathcal{M}([\mathbf{E}_{\mathrm{2D}}\Vert\mathbf{E}_{\mathrm{3D}}])$.
This augments image-plane position with information about the reconstructed 3D scene.
Deformable attention uses the geometry-conditioned features $\widetilde{\mathbf{F}}$ and positional embeddings $\widetilde{\mathbf{E}}$ when updating the object queries.
The resulting queries therefore encode both visual evidence and estimated 3D scene information.
We refer to this process as \emph{geometric query lifting} because it extends the query representation beyond the image plane while preserving the detector's existing 2D components.

\begin{table*}[t]
    \centering
    \caption{Cumulative contributions of geometric query lifting (GQL), anchored 3D localization (A3DL), and dual-resolution temporal memory (DRTM). GQL and A3DL jointly form world-grounded queries. Components are enabled cumulatively from top to bottom.}
    \label{tab:main-ablation}
    \setlength{\tabcolsep}{5pt}
    \renewcommand{\arraystretch}{1.08}
    \begin{tabular}{l c c c c c c c c}
        \toprule
        & \multicolumn{2}{c}{World-Grounded Queries} & \multicolumn{1}{c}{Memory} & \multicolumn{5}{c}{Metrics} \\
        \cmidrule(lr){2-3} \cmidrule(lr){4-4} \cmidrule(lr){5-9}
        Variant & GQL & A3DL & DRTM & HOTA $\uparrow$ & AssA $\uparrow$ & DetA $\uparrow$ & IDF1 $\uparrow$ & MOTA $\uparrow$ \\
        \midrule
        Base & \xmark & \xmark & \xmark & 63.0 & 53.2 & 75.0 & 66.6 & 84.7 \\
        + GQL & \cmark & \xmark & \xmark & 64.0 & 54.3 & 75.9 & 67.6 & 85.5 \\
        + A3DL & \cmark & \cmark & \xmark & 64.4 & 54.9 & \colorbox{green!30}{\textbf{76.1}} & 67.8 & 85.5 \\
        + DRTM & \cmark & \cmark & \cmark & \colorbox{green!30}{\textbf{65.6}} & \colorbox{green!30}{\textbf{56.9}} & \colorbox{green!30}{\textbf{76.1}} & \colorbox{green!30}{\textbf{70.3}} & \colorbox{green!30}{\textbf{85.9}} \\
        \bottomrule
    \end{tabular}
\end{table*}

\subsubsection{Anchored 3D Localization}
In Deformable DETR, each object query is associated with a learned 2D reference point that guides its localization.
Bounding-box prediction proceeds through iterative refinements of this reference point across the transformer decoder, resulting in the final 2D localization.

A direct extension to 3D would learn an independent 3D reference point from scratch.
Yet the backbone and detector have already acquired strong localization priors through large-scale 2D pretraining.
Learning an isolated 3D reference would leave these priors underused and separate 3D localization from the detector's established spatial representation.
We therefore extend the motivation behind geometric query lifting to the reference points themselves by lifting the detector's existing 2D reference point $\mathbf{r}_{\mathrm{2D}}$ into a corresponding 3D reference point $\mathbf{r}_{\mathrm{3D}}$ using the aligned scene geometry.
Given $\mathbf{r}_{\mathrm{3D}}$, an additional lightweight 3D prediction head estimates a residual from the corresponding query embedding $\mathbf{q}$ and refines the reference point to predict the object's 3D center.
This anchor-relative formulation builds on the detector's pretrained 2D localization mechanism instead of requiring absolute 3D positions to be inferred independently.
After the detector performs its standard matching, valid 3D labels supervise the corresponding predicted centers through an additional loss $\mathcal{L}_{\ell_1}^{\mathrm{3D}}$.
During training, this term is added to the existing detection and identity objectives.

Anchored 3D localization complements geometric query lifting: lifting makes dense geometric evidence available during query formation, while localization encourages matched queries to retain object-specific 3D position information.

\subsection{Dual-Resolution Temporal Memory}
Transformer-based end-to-end trackers incur increasing computational cost as their identity models process and attend to a growing history of query representations.
Their temporal context is therefore typically restricted, making older identity evidence inaccessible across extended occlusions.
\modelname addresses this limitation with \emph{dual-resolution temporal memory}, which extends the accessible temporal horizon within a fixed context budget.
Specifically, we divide the fixed identity-decoder context into complementary short-term and long-term regions.
The dense short-term region preserves the latest evolution of the scene, supporting consistent association through brief occlusions, missed detections, and rapid motion.
The sparsely sampled long-term region allows the decoder to revisit identity evidence beyond its original consecutive-frame horizon, supporting trajectory continuity across extended disappearances without increasing the number of processed representations.
To make the identity model robust to these different temporal densities, we also vary the temporal sampling rate during training.
This prepares the decoder to associate evidence across the short-term and long-term gaps selected by dual-resolution memory.

Formally, given a context budget of $M$ observations, we allocate approximately $M/2$ slots to the most recent consecutive frames and the remaining $M/2$ slots to earlier frames sampled with stride $S$.
Because the trajectory buffer stores the detector's final query embeddings, this extended context preserves world-grounded evidence across longer gaps without modifying the detector outputs or association architecture.
This extension also does not require additional training time.

\section{Experiments}
\label{sec:experiments}
\subsection{Datasets and Metrics}
\label{sec:datasets-and-metrics}

\subsubsection{Datasets}
\label{sec:datasets}

\noindent\textbf{DanceTrack.}
DanceTrack contains 100 videos, primarily of group dancing, in which people with similar appearances undergo complex motion, articulation, crossovers, and occlusions~\cite{sun2022dancetrack}.
Because appearance is often weakly discriminative, the benchmark places particular emphasis on motion-aware association.

\noindent\textbf{SportsMOT.}
SportsMOT comprises 240 sequences from basketball, volleyball, and football, totaling more than 150{,}000 frames and 1.6 million annotated bounding boxes~\cite{cui2023sportsmot}.
Its fast, variable-speed player motion makes association challenging, while similar player appearances limit the reliability of visual identity cues.

\noindent\textbf{BFT.}
The Bird Flock Tracking (BFT) benchmark contains 106 clips spanning 22 bird species and 14 natural and human-made scene types~\cite{zheng2024nettrack}.
It targets highly dynamic open-world tracking, with rapid motion, deformation, frequent occlusion, and visually similar objects in dense flocks.

\subsubsection{Metrics}
\label{sec:metrics}

We report Higher Order Tracking Accuracy (HOTA), Identification F1 (IDF1), and Multiple Object Tracking Accuracy (MOTA), with higher values indicating better performance.

\noindent\textbf{HOTA}~\cite{luiten2021hota}.
At each localization threshold, HOTA computes the geometric mean of detection accuracy (DetA) and association accuracy (AssA), and the final score averages over thresholds.
Its explicit combination of detection and association provides a balanced summary of these two aspects of tracking performance.

\noindent\textbf{IDF1}~\cite{ristani2016performance}.
IDF1 is the harmonic mean of identification precision and identification recall after a global one-to-one assignment between predicted and ground-truth identities.
It therefore emphasizes identity preservation over the sequence and is particularly informative about association consistency.

\noindent\textbf{MOTA}~\cite{bernardin2008clear}.
MOTA combines false negatives, false positives, and identity mismatches into an error count normalized by the number of ground-truth detections.
Because false positives and missed detections contribute directly to this count, MOTA tends to emphasize detection coverage more than long-term identity preservation.

\subsection{Implementation Details}
\label{sec:implementation-details}

To make our comparisons with prior tracking-by-query methods as direct as possible, we use the same established detector family~\cite{gao2025multiple,gao2023memotr,yu2023motrv3,segu2025samba}: Deformable DETR~\cite{zhu2021deformable} with a ResNet-50 backbone~\cite{he2016deep}.
Following the initialization practice of prior works~\cite{gao2023memotr,yan2025comot,yu2023motrv3}, we initialize the detector from COCO-pretrained weights.
Together, these choices keep our experimental setting aligned with prior work.

We follow the training setup of MOTIP~\cite{gao2025multiple} and train with AdamW~\cite{loshchilov2019decoupled} using an initial learning rate of $1\times10^{-4}$ and a weight decay of $5\times10^{-4}$.
Training also includes random resizing, cropping, horizontal flipping, trajectory occlusion, and identity switching.
We train each ablation configuration five times and report the mean result to reduce sensitivity to run-to-run variation.

The added geometric components are lightweight, and \modelname retains a runtime comparable to MOTIP, processing video at 12 FPS in FP32 and 20 FPS in FP16.

\subsection{Ablations}
\label{sec:ablations}

\subsubsection{Main Components}
\label{sec:ablation-core-components}

\noindent\textbf{Impact of World-Grounded Queries and Temporal Memory.}
\Cref{tab:main-ablation} evaluates the cumulative contribution of geometric query lifting, anchored 3D localization, and dual-resolution temporal memory, with the first two components jointly forming world-grounded queries.
Geometric query lifting improves every reported metric over the base model, including a 1.0-point gain in HOTA.
The comparable gains in association and detection accuracy indicate that incorporating estimated 3D position information during query formation benefits both aspects of tracking.

Adding anchored 3D localization on top of geometric query lifting yields a further 0.4-point improvement in HOTA, with a larger gain in association accuracy than in detection accuracy.
This result shows that training matched queries to localize objects in 3D through anchored refinement helps the end-to-end tracker resolve association ambiguities caused by occlusion, rather than merely improving detection.

Finally, dual-resolution temporal memory improves HOTA by another 1.2 points and produces clear association gains while detection accuracy remains unchanged.
These gains highlight the importance of extending the model context from short-term observations to long-term temporal information, with stable 3D spatial cues supporting correspondences over longer time horizons.

\noindent\textbf{Interaction Between 3D Awareness and Temporal Memory.}
Having established the importance of temporal memory, we next ask how its effectiveness changes when the tracker has access to 3D information.
\Cref{tab:memory-ablation} addresses this question by measuring the gain from adding dual-resolution temporal memory to two otherwise corresponding models: one that reasons only from 2D information and one equipped with world-grounded queries.
For the 2D-aware model, dual-resolution temporal memory improves HOTA by 0.8 points and IDF1 by 1.5 points, while MOTA remains unchanged.
This result confirms that extending the temporal horizon is beneficial even without explicit 3D information.
For the 3D-aware model, the same memory mechanism produces larger gains across all reported metrics, improving HOTA by 1.2 points, IDF1 by 2.5 points, and MOTA by 0.4 points.
Thus, temporal memory benefits both models, but its stronger effect in the 3D-aware setting further highlights the synergy between 3D information and extended temporal context.
Overall, world-grounded representations provide stable spatial evidence, while dual-resolution temporal memory makes this evidence available for identity reasoning across longer gaps.
\begin{table}[t]
    \centering
    \caption{Interaction between 3D awareness and dual-resolution temporal memory (DRTM). Each cell reports performance with basic memory, with DRTM, and the corresponding change in parentheses. Highlighted values indicate the larger DRTM gain for each metric.}
    \label{tab:memory-ablation}
    \setlength{\tabcolsep}{3.5pt}
    \renewcommand{\arraystretch}{1.12}
    \begin{tabular}{l c c c}
        \toprule
        & \multicolumn{3}{c}{Basic Memory $\rightarrow$ DRTM (Gain)} \\
        \cmidrule(lr){2-4}
        Model & HOTA $\uparrow$ & IDF1 $\uparrow$ & MOTA $\uparrow$ \\
        \midrule
        2D-aware
        & \shortstack{63.0 $\rightarrow$ 63.8 \\ {\small $(+0.8)$}}
        & \shortstack{66.6 $\rightarrow$ 68.1 \\ {\small $(+1.5)$}}
        & \shortstack{84.7 $\rightarrow$ 84.7 \\ {\small $(+0.0)$}} \\
        \addlinespace[3pt]
        3D-aware
        & \shortstack{64.4 $\rightarrow$ 65.6 \\ {\small (\colorbox{green!30}{$\mathbf{+1.2}$})}}
        & \shortstack{67.8 $\rightarrow$ 70.3 \\ {\small (\colorbox{green!30}{$\mathbf{+2.5}$})}}
        & \shortstack{85.5 $\rightarrow$ 85.9 \\ {\small (\colorbox{green!30}{$\mathbf{+0.4}$})}} \\
        \bottomrule
    \end{tabular}
\end{table}

\begin{table}[t]
    \centering
    \caption{Direct versus anchored 3D localization. D3DL denotes direct 3D localization, while A3DL denotes anchored 3D localization.}
    \label{tab:model-structure-ablation}
    \setlength{\tabcolsep}{3pt}
    \renewcommand{\arraystretch}{1.08}
    \resizebox{\columnwidth}{!}{%
    \begin{tabular}{l c c c c c}
        \toprule
        Model Variant & HOTA $\uparrow$ & AssA $\uparrow$ & DetA $\uparrow$ & IDF1 $\uparrow$ & MOTA $\uparrow$ \\
        \midrule
        Base & 64.0 & 54.3 & 75.9 & 67.6 & \colorbox{green!30}{\textbf{85.5}} \\
        \midrule
        D3DL (Naive) & 63.8 & 54.2 & 75.7 & 67.4 & 85.4 \\
        A3DL (Ours) & \colorbox{green!30}{\textbf{64.4}} & \colorbox{green!30}{\textbf{54.9}} & \colorbox{green!30}{\textbf{76.1}} & \colorbox{green!30}{\textbf{67.8}} & \colorbox{green!30}{\textbf{85.5}} \\
        \bottomrule
    \end{tabular}
    }
\end{table}

\subsubsection{How to Embed 3D Cues into a 2D Model?}
\label{sec:ablation-3d-localization}

\noindent\textbf{Direct vs. Anchored 3D Localization.}
\Cref{tab:model-structure-ablation} compares two models trained with the same 3D supervision but different 3D localization paradigms.
Direct 3D Localization (\textit{Naive}) adds a 3D prediction head in parallel with the existing 2D head and learns an independent reference point in 3D space.
Anchored 3D Localization (\textit{Ours}) instead lifts the detector's existing 2D reference point into 3D and predicts a refinement relative to this anchor.
The direct version reduces HOTA from 64.0 to 63.8 and produces small declines across all reported metrics.
In contrast, our anchored version outperforms the direct version across all metrics, improving HOTA by 0.6 points and exceeding the baseline by 0.4 points.
By building on the pretrained 2D localization capability, our version avoids learning 3D localization entirely from scratch.
This coupling allows 2D and 3D localization to shape a shared representation, making the additional 3D task easier to learn while retaining useful 2D information.
These results show that although 3D supervision provides valuable geometric information, its benefit depends critically on how it is integrated into the model, supporting the refinement design used in our anchored 3D localization.

\noindent\textbf{3D Foundation Features vs. World-Grounded Queries.}
A straightforward way to introduce 3D cues into a 2D tracker is to inject features extracted by a pretrained 3D foundation model.
We use this strategy as a baseline to our world-grounded queries by extracting \textit{3D Foundation Features} from Depth Anything 3~\cite{lin2025depth} and fusing them into the tracker during training.
As shown in \cref{tab:3d-information-usage-ablation}, these features improve HOTA only marginally, from 63.0 to 63.1, while leaving association accuracy and IDF1 unchanged or slightly reduced.
Although the injected features contain strong 3D cues, they are not assigned an explicit role in query formation or the tracking objective.
Moreover, the foundation model is trained for geometry estimation rather than tracking, so its features must be adapted to the tracking domain and objectives before the model can fully capitalize on them.
The limited gains suggest that tracking supervision through generic feature fusion alone does not provide a sufficiently effective interface for this adaptation.

World-grounded queries instead introduce 3D information through geometric query lifting and anchored 3D localization, allowing dense scene geometry to condition query formation while explicitly encouraging matched queries to encode object-level 3D locations.
This structured formulation improves HOTA by 1.4 points over the baseline and yields gains in both association and detection accuracy.
Overall, these results show that incorporating pretrained 3D cues into a 2D tracker is not straightforward and requires design decisions that align geometric information with the model's query representations and objectives.
\begin{table}[t]
    \centering
    \caption{3D foundation features versus the world-grounded query formulation. The latter combines geometric query lifting with anchored 3D localization.}
    \label{tab:3d-information-usage-ablation}
    \setlength{\tabcolsep}{3pt}
    \renewcommand{\arraystretch}{1.08}
    \resizebox{\columnwidth}{!}{%
    \begin{tabular}{l c c c c c}
        \toprule
        3D Cues & HOTA $\uparrow$ & AssA $\uparrow$ & DetA $\uparrow$ & IDF1 $\uparrow$ & MOTA $\uparrow$ \\
        \midrule
        Base & 63.0 & 53.2 & 75.0 & 66.6 & 84.7 \\
        \midrule
        3D Foundation Features & 63.1 & 53.1 & 75.3 & 66.6 & 84.9 \\
        WGQ (Ours) & \colorbox{green!30}{\textbf{64.4}} & \colorbox{green!30}{\textbf{54.9}} & \colorbox{green!30}{\textbf{76.1}} & \colorbox{green!30}{\textbf{67.8}} & \colorbox{green!30}{\textbf{85.5}} \\
        \bottomrule
    \end{tabular}
    }
\end{table}

\begin{table}[t]
    \centering
    \caption{Benchmark results on the DanceTrack test set.}
    \label{tab:benchmark-dancetrack}
    \setlength{\tabcolsep}{3pt}
    \renewcommand{\arraystretch}{1.08}
    \resizebox{\columnwidth}{!}{%
    \begin{tabular}{l c c c c c}
        \toprule
        \multicolumn{6}{c}{DanceTrack} \\
        \midrule
        Method & HOTA $\uparrow$ & AssA $\uparrow$ & DetA $\uparrow$ & IDF1 $\uparrow$ & MOTA $\uparrow$ \\
        \midrule
        \multicolumn{6}{l}{\textit{Tracking-by-Detection}} \\
        \cmidrule(lr){1-6}
        FairMOT~\cite{zhang2021fairmot}         & 39.7 & 23.8 & 66.7 & 40.8 & 82.2 \\
        CenterTrack~\cite{zhou2020tracking}     & 41.8 & 22.6 & 78.1 & 35.7 & 86.8 \\
        TraDeS~\cite{wu2021trades}              & 43.3 & 25.4 & 74.5 & 41.2 & 86.2 \\
        DeepSORT~\cite{wojke2017simple}         & 45.6 & 29.7 & 71.0 & 47.9 & 87.8 \\
        ByteTrack~\cite{zhang2022bytetrack}     & 47.7 & 32.1 & 71.0 & 53.9 & 89.6 \\
        SORT~\cite{bewley2016simple}            & 47.9 & 31.2 & 72.0 & 50.8 & 91.8 \\
        GTR~\cite{zhou2022global}               & 48.0 & 31.9 & 72.5 & 50.3 & 84.7 \\
        QDTrack~\cite{pang2021quasidense}       & 54.2 & 36.8 & 80.1 & 50.4 & 87.7 \\
        OC-SORT~\cite{cao2023observationcentric} & 55.1 & 38.3 & 80.3 & 54.6 & 92.0 \\
        StrongSORT~\cite{du2023strongsort}      & 55.6 & 38.6 & 80.7 & 55.2 & 91.1 \\
        SparseTrack~\cite{liu2025sparsetrack}   & 55.7 & 39.3 & 79.2 & 58.1 & 91.3 \\
        C-BIoU~\cite{yang2023hard}              & 60.6 & 45.4 & 81.3 & 61.6 & 91.6 \\
        Hybrid-SORT~\cite{yang2024hybridsort}   & 62.2 & --   & --   & 63.0 & 91.6 \\
        DiffMOT~\cite{lv2024diffmot}            & 62.3 & 47.2 & \colorbox{green!30}{\textbf{82.5}} & 63.0 & 92.8 \\
        TrackTrack~\cite{shim2025focusing}      & 66.5 & 52.9 & --   & 67.8 & \colorbox{green!30}{\textbf{93.6}} \\
        \midrule
        \multicolumn{6}{l}{\textit{End-to-End}} \\
        \cmidrule(lr){1-6}
        TransTrack~\cite{sun2020transtrack} & 45.5 & 27.5 & 75.9 & 45.2 & 88.4 \\
        MOTR~\cite{zeng2022motr}            & 54.2 & 40.2 & 73.5 & 51.5 & 79.7 \\
        MeMOTR~\cite{gao2023memotr}          & 63.4 & 52.3 & 77.0 & 65.5 & 85.4 \\
        CO-MOT~\cite{yan2025comot}           & 65.3 & 53.5 & 80.1 & 66.5 & 89.3 \\
        LA-MOTR~\cite{wang2025lamotr}        & 66.5 & 53.9 & 80.4 & 69.7 & 92.5 \\
        SambaMOTR~\cite{segu2025samba}       & 67.2 & 57.5 & 78.8 & 70.5 & 88.1 \\
        MOTIP~\cite{gao2025multiple}         & 69.6 & 60.4 & 80.4 & 74.7 & 90.6 \\
        \midrule
        \modelnamewithplanet{} (Ours) & \colorbox{green!30}{\textbf{72.1}} & \colorbox{green!30}{\textbf{63.9}} & 81.4 & \colorbox{green!30}{\textbf{78.0}} & 91.6 \\
        \bottomrule
    \end{tabular}
    }
\end{table}

\subsection{Benchmark Results}
\label{sec:benchmark-results}

Following the evaluation protocol of MOTIP~\cite{gao2025multiple}, we evaluate \modelname on the same three benchmarks: DanceTrack, SportsMOT, and BFT.
We also adopt MOTIP's dataset-specific training scheme, training a model exclusively on each corresponding benchmark without pretraining or joint training on related tracking or detection datasets.

Many multi-object tracking methods instead pretrain on additional datasets or augment benchmark training data with external datasets~\cite{zhang2022bytetrack,cao2023observationcentric,yan2025comot,yu2023motrv3,zhang2023motrv2,lv2024diffmot}.
A common choice is CrowdHuman~\cite{shao2018crowdhuman}, whose static images can be transformed into pseudo-video sequences.
MOTIP argues that this practice can produce models that generalize poorly across tracking settings and complicate fair comparison because methods may benefit from different sources and amounts of additional supervision~\cite{gao2025multiple}.
For further discussion of this evaluation choice, we refer to the MOTIP paper~\cite{gao2025multiple}.

Consequently, our reported results do not benefit from external training data, whereas several competing methods are trained with such additional supervision.
Despite this more restrictive protocol, \modelname reaches state-of-the-art performance on all three benchmarks and outperforms methods both with and without external training data.

\noindent\textbf{DanceTrack.}
DanceTrack evaluates association under complex motion, frequent crossovers and occlusions, and limited appearance diversity among the tracked dancers.
In \Cref{tab:benchmark-dancetrack}, \modelname achieves 72.1 HOTA, exceeding MOTIP by 2.5 points, and raises IDF1 from 74.7 to 78.0 while improving every reported metric over this strongest end-to-end baseline.
The gains across detection and association demonstrate the effectiveness of the complete model under ambiguous appearance and motion patterns.

\noindent\textbf{SportsMOT.}
SportsMOT tests tracking under fast, variable player motion and similar appearances across diverse sports scenes.
In \Cref{tab:benchmark-sportsmot}, \modelname achieves state-of-the-art performance across the reported metrics, reaching 73.7 HOTA, 1.1 points above MOTIP, while raising DetA from 83.5 to 86.0 and MOTA from 92.4 to 95.2.
These results show that the method remains effective in large-scale sequences with rapid and irregular motion.

\noindent\textbf{BFT.}
BFT evaluates the method in dense bird flocks characterized by rapid motion, deformation, frequent occlusion, and visually similar individuals.
In \Cref{tab:benchmark-bft}, \modelname achieves state-of-the-art performance across all five metrics, including 72.6 HOTA, exceeding the previous state of the art by 2.1 points.
The gains across detection and association demonstrate the effectiveness of world-grounded queries and dual-resolution temporal memory in highly dynamic, densely interacting scenes.

\begin{table}[t]
    \centering
    \caption{Benchmark results on the SportsMOT test set.}
    \label{tab:benchmark-sportsmot}
    \setlength{\tabcolsep}{3pt}
    \renewcommand{\arraystretch}{1.08}
    \resizebox{\columnwidth}{!}{%
    \begin{tabular}{l c c c c c}
        \toprule
        \multicolumn{6}{c}{SportsMOT} \\
        \midrule
        Method & HOTA $\uparrow$ & AssA $\uparrow$ & DetA $\uparrow$ & IDF1 $\uparrow$ & MOTA $\uparrow$ \\
        \midrule
        \multicolumn{6}{l}{\textit{Tracking-by-Detection}} \\
        \cmidrule(lr){1-6}
        FairMOT~\cite{zhang2021fairmot}         & 49.3 & 34.7 & 70.2 & 53.5 & 86.4 \\
        GTR~\cite{zhou2022global}               & 54.5 & 45.9 & 64.8 & 55.8 & 67.9 \\
        QDTrack~\cite{pang2021quasidense}       & 60.4 & 47.2 & 77.5 & 62.3 & 90.1 \\
        CenterTrack~\cite{zhou2020tracking}     & 62.7 & 48.0 & 82.1 & 60.0 & 90.8 \\
        ByteTrack~\cite{zhang2022bytetrack}     & 62.8 & 51.2 & 77.1 & 69.8 & 94.1 \\
        OC-SORT~\cite{cao2023observationcentric} & 68.1 & 54.8 & 84.8 & 68.0 & 93.4 \\
        BoT-SORT~\cite{aharon2022botsort}       & 68.7 & 55.9 & 84.4 & 70.0 & 94.5 \\
        DiffMOT~\cite{lv2024diffmot}            & 72.1 & 60.5 & \colorbox{green!30}{\textbf{86.0}} & 72.8 & 94.5 \\
        \midrule
        \multicolumn{6}{l}{\textit{End-to-End}} \\
        \cmidrule(lr){1-6}
        MeMOTR~\cite{gao2023memotr}       & 68.8 & 57.8 & 82.0 & 69.9 & 90.2 \\
        TransTrack~\cite{sun2020transtrack} & 68.9 & 57.5 & 82.7 & 71.5 & 92.6 \\
        LA-MOTR~\cite{wang2025lamotr}     & 69.5 & 57.9 & 84.4 & 70.0 & 93.5 \\
        SambaMOTR~\cite{segu2025samba}   & 69.8 & 59.4 & 82.2 & 71.9 & 90.3 \\
        MOTIP~\cite{gao2025multiple}     & 72.6 & \colorbox{green!30}{\textbf{63.2}} & 83.5 & 77.1 & 92.4 \\
        \midrule
        \modelnamewithplanet{} (Ours) & \colorbox{green!30}{\textbf{73.7}} & \colorbox{green!30}{\textbf{63.2}} & \colorbox{green!30}{\textbf{86.0}} & \colorbox{green!30}{\textbf{78.7}} & \colorbox{green!30}{\textbf{95.2}} \\
        \bottomrule
    \end{tabular}
    }
\end{table}

\begin{table}[t]
    \centering
    \caption{Benchmark results on the BFT test set.}
    \label{tab:benchmark-bft}
    \setlength{\tabcolsep}{3pt}
    \renewcommand{\arraystretch}{1.08}
    \resizebox{\columnwidth}{!}{%
    \begin{tabular}{l c c c c c}
        \toprule
        \multicolumn{6}{c}{BFT} \\
        \midrule
        Method & HOTA $\uparrow$ & AssA $\uparrow$ & DetA $\uparrow$ & IDF1 $\uparrow$ & MOTA $\uparrow$ \\
        \midrule
        \multicolumn{6}{l}{\textit{Tracking-by-Detection}} \\
        \cmidrule(lr){1-6}
        JDE~\cite{wang2020towards}               & 30.7 & 23.4 & 40.9 & 37.4 & 35.4 \\
        CSTrack~\cite{liang2022rethinking}       & 33.2 & 23.7 & 47.0 & 34.5 & 46.7 \\
        FairMOT~\cite{zhang2021fairmot}          & 40.2 & 28.2 & 53.3 & 41.8 & 56.0 \\
        SORT~\cite{bewley2016simple}             & 61.2 & 62.3 & 60.6 & 77.2 & 75.5 \\
        ByteTrack~\cite{zhang2022bytetrack}      & 62.5 & 64.1 & 61.2 & 82.3 & 77.2 \\
        CenterTrack~\cite{zhou2020tracking}      & 65.0 & 54.0 & 58.5 & 61.0 & 60.2 \\
        OC-SORT~\cite{cao2023observationcentric} & 66.8 & 68.7 & 65.4 & 79.3 & 77.1 \\
        \midrule
        \multicolumn{6}{l}{\textit{End-to-End}} \\
        \cmidrule(lr){1-6}
        TransCenter~\cite{xu2021transcenter}       & 60.0 & 61.1 & 66.0 & 72.4 & 74.1 \\
        TransTrack~\cite{sun2020transtrack}        & 62.1 & 60.3 & 64.2 & 71.4 & 71.4 \\
        TrackFormer~\cite{meinhardt2022trackformer} & 63.3 & 61.1 & 66.0 & 72.4 & 74.1 \\
        SambaMOTR~\cite{segu2025samba}             & 69.6 & 73.6 & 66.0 & 81.9 & 72.0 \\
        MOTIP~\cite{gao2025multiple}               & 70.5 & 71.8 & 69.6 & 82.1 & 77.1 \\
        \midrule
        \modelnamewithplanet{} (Ours) & \colorbox{green!30}{\textbf{72.6}} & \colorbox{green!30}{\textbf{73.9}} & \colorbox{green!30}{\textbf{71.6}} & \colorbox{green!30}{\textbf{84.6}} & \colorbox{green!30}{\textbf{80.4}} \\
        \bottomrule
    \end{tabular}
    }
\end{table}

\section{Conclusion}
\label{sec:conclusion}

We have introduced \modelname, an end-to-end multi-object tracking framework that moves beyond image-plane reasoning by grounding object queries in reconstructed 3D scene geometry and retaining this information across longer temporal horizons.
Through the careful design of geometric query lifting and anchored 3D localization, \modelname incorporates 3D information directly into a 2D tracker.
Our ablation studies validate the contribution of each component and show that world-grounded queries and dual-resolution temporal memory provide complementary gains.
Our benchmark results demonstrate state-of-the-art performance across diverse tracking scenarios, supporting the generality of the approach.
We believe this work represents a step toward using 3D knowledge to improve tasks traditionally formulated in 2D.
We hope \modelname encourages future research to look beyond image-plane evidence and pursue richer forms of 3D reasoning as a central direction for addressing the ambiguities that continue to limit tracking and visual understanding.

\clearpage
{
    \small
    \bibliographystyle{ieeenat_fullname}
    \bibliography{main}
}

\end{document}